\documentclass[10pt,twocolumn,letterpaper]{article}

\usepackage[pagenumbers]{cvpr} % To force page numbers, e.g. for an arXiv version

\definecolor{tabfirst}{rgb}{1, 0.7, 0.7} % red
\definecolor{tabsecond}{rgb}{1, 0.85, 0.7} % orange
\definecolor{tabthird}{rgb}{1, 1, 0.7} % yellow

\newcommand\customparagraph[1]{\vspace{0.4em}\noindent\textbf{#1.}}

\newcommand{\scenename}[1]{\textit{#1}}
\definecolor{cvprblue}{rgb}{0.21,0.49,0.74}
\usepackage[pagebackref,breaklinks,colorlinks,allcolors=cvprblue]{hyperref}

\usepackage{bm}
\usepackage{multirow}
\usepackage{multicol}
\usepackage[dvipsnames]{xcolor}
\usepackage{colortbl}
\usepackage{booktabs}

\def\paperID{xxxxx} % *** Enter the Paper ID here
\def\confName{CVPR}
\def\confYear{2025}

\title{Pushing Rendering Boundaries: Hard Gaussian Splatting}

\author{Qingshan Xu$^{1}$\quad Jiequan Cui$^{1}$\quad Xuanyu Yi$^{1}$\quad Yuxuan Wang$^{1}$\quad Yuan Zhou$^{1}$ \\ Yew-Soon Ong$^{1,2}$\quad Hanwang Zhang$^{1}$\vspace{8pt} \\
$^{1}$ Nanyang Technological University \qquad $^{2}$ A*STAR, Singapore
}

\begin{document}

\twocolumn[{
\renewcommand\twocolumn[1][]{#1}
\maketitle
\centering
\vspace{-0.5cm}
 \includegraphics[width=\linewidth]{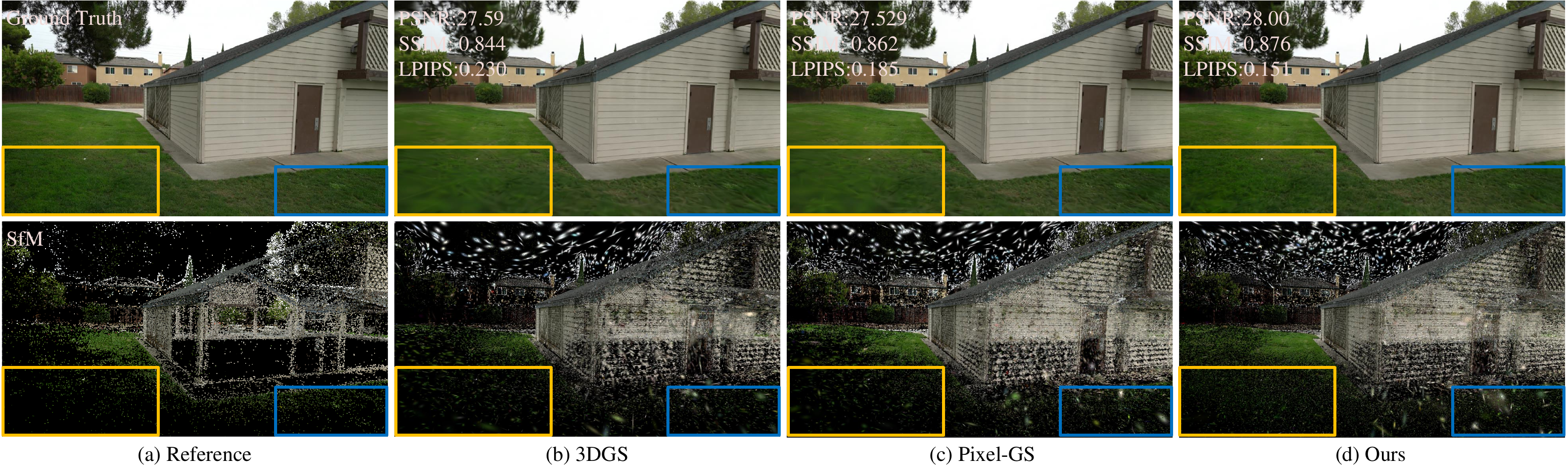}

 \captionsetup{type=figure}
\caption{\textbf{Illustration of the relationship between rendering and Gaussians distribution.} We show that 3DGS~\cite{kerbl20233d} and Pixel-GS~\cite{zhang2024pixel} are unable to grow Gaussians in some challenging areas, leading to artifacts in these areas. In contrast, our method effectively grows \emph{hard} Gaussians in these challenging areas to recover 3D scenes, thus achieving better novel view synthesis results.   
}
\vspace{0.45cm}
\label{fig:teaser}
}]

\maketitle
\begin{abstract}
3D Gaussian Splatting (3DGS) has demonstrated impressive Novel View Synthesis (NVS) results in a real-time rendering manner. During training, it relies heavily on the average magnitude of view-space positional gradients to grow Gaussians to reduce rendering loss. However, this average operation smooths the positional gradients from different viewpoints and rendering errors from different pixels, hindering the growth and optimization of many defective Gaussians. This leads to strong spurious artifacts in some areas. To address this problem, we propose Hard Gaussian Splatting, dubbed HGS, which considers multi-view significant positional gradients and rendering errors to grow hard Gaussians that fill the gaps of classical Gaussian Splatting on 3D scenes, thus achieving superior NVS results. In detail, we present positional gradient driven HGS, which leverages multi-view significant positional gradients to uncover hard Gaussians. Moreover, we propose rendering error guided HGS, which identifies noticeable pixel rendering errors and potentially over-large Gaussians to jointly mine hard Gaussians. By growing and optimizing these hard Gaussians, our method helps to resolve blurring and needle-like artifacts. Experiments on various datasets demonstrate that our method achieves state-of-the-art rendering quality while maintaining real-time efficiency. Our code will be available at \url{https://github.com/GhiXu/HGS}.
\end{abstract}    
\section{Introduction}
\label{sec:intro}

Novel View Synthesis (NVS) is a fundamental task in computer vision and graphics, with its wide applications in virtual reality, robotics and media generation. In the past few years, Neural Radiance Field (NeRF)~\cite{mildenhall2021nerf} has made significant advances in NVS. It adopts neural implicit representations to model 3D scenes via volume rendering~\cite{max1995optical}. Generally, NeRF works~\cite{barron2021mip,barron2022mip,muller2022instant} require ray marching for volume rendering, which compromises on efficiency. Recently, 3D Gaussian Splatting (3DGS)~\cite{kerbl20233d} combines an point-based explicit representation, 3D Gaussian, with GPU-friendly differentiable rasterization for volume rendering, achieving impressive NVS results in a real-time manner.

\begin{figure}[t]
    \centering
    \includegraphics[width=\linewidth]{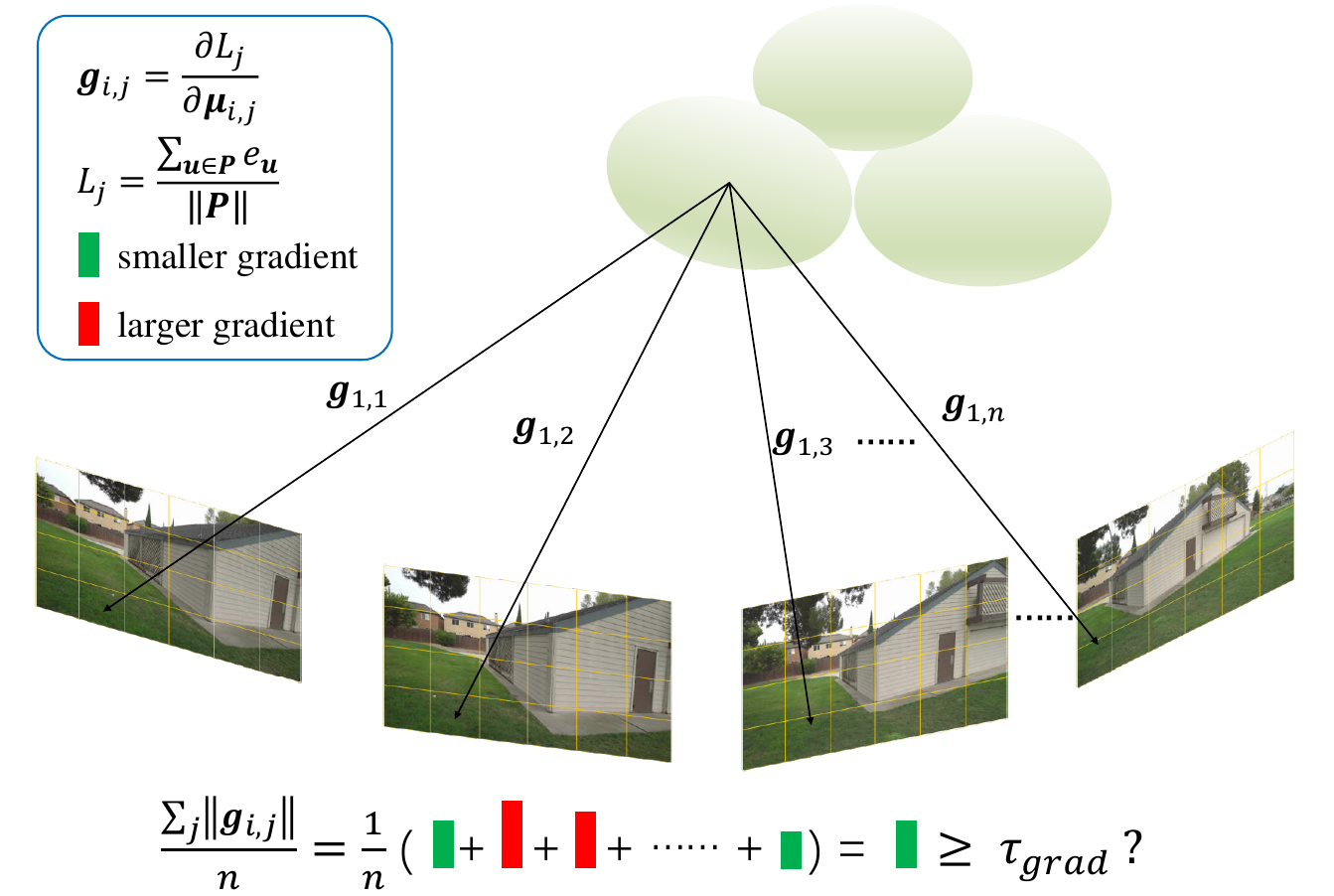}
    \caption{\textbf{Illustration of the Gaussian growing criterion in 3DGS~\cite{kerbl20233d}.} $\boldsymbol{g}_{i,j}$ denotes the positional gradient of Gaussian $\mathcal{G}_i$ under viewpoint $j$. $L_j$ denotes the rendering loss under viewpoint $j$, which is computed by averaging the rendering errors $\{e_{\boldsymbol{u}}\}$ of all pixels $\bm{P}$. Larger gradients and rendering errors will be smoothed by the average operation.}
    \label{fig:motivation}
\end{figure}

3DGS initializes a set of 3D Gassuains from \emph{sparse} point clouds produced by Structure from Motion (SfM)~\cite{snavely2006photo} to represent a scene, and gradually populates empty areas by growing existing Gaussians (the bottom row of Figures~\ref{fig:teaser}(a) and (b)). To select good candidates for growing, 3DGS checks if the \emph{average} magnitude of view-space positional gradients of each Gaussian in a growth interval is larger than a threshold, as illustrated in Figure~\ref{fig:motivation}. This criterion encourages Gaussians to gradually \emph{densify} and reconstruct the entire 3D scene. 

However, strong spurious artifacts remain in some challenging regions, such as areas with repetitive textures and few observations, as shown in the yellow and blue boxes in the top row of Figures~\ref{fig:teaser}(b) and (c). By observing the corresponding Gaussians distribution in the bottom row of Figures~\ref{fig:teaser}(b) and (c), we find that there are very few Gaussians in these challenging areas. This demonstrates that the growing criterion, in fact, cannot \emph{adequately} select suitable candidates from existing Gaussians for growing, thus limiting the NVS performance of 3DGS. We define these candidates as \emph{hard Gaussians}. Intuitively, the average operation used in the growing criterion makes it unable to reflect the larger positional gradients of some viewpoints, as shown in Figure~\ref{fig:motivation}. This leads to rendering inconsistency across views. Moreover, the rendering loss used to optimize 3DGS indicates the rendering quality of the entire image but fails to indicate the quality in local image regions. Therefore, the average magnitude of positional gradients cannot reflect noticeable rendering errors of certain local image regions, especially when these regions are only observed by few viewpoints. These deficiencies 
% of the growing criterion 
prevent 3DGS from growing \emph{hard} Gaussians to achieve better NVS results.   

To address these issues, we propose Hard Gaussian Splatting, dubbed HGS, to fill the gaps of classical Gaussian Splatting on 3D scenes. Our key insight is to uncover hard Gaussians from multi-view significant positional gradients and rendering errors, and grow these Gaussians for subsequent optimization. Specifically, besides the original Gaussian growing criterion, we first present positional gradient driven HGS (Sec.~\ref{sec:MH-PG}). This strategy captures $k$ largest positional gradients for each Gaussian in the growth interval. If the minimum of these positional gradients exceeds a threshold, its corresponding Gaussian is deemed as a hard Gaussian and will grow. This strategy exploits cross-view significant positional gradients to uncover hard Gaussians, thus alleviating the rendering inconsistency across views. 

In addition, we propose rendering error guided HGS (Sec.~\ref{sec:MH-RE}). This strategy first identifies potentially over-large Gaussians that may cause blurring artifacts in the current training image. These Gaussians are then projected into the image to calculate pixel rendering errors. If noticeable pixel rendering errors are detected from multiple viewpoints within the growth interval, the corresponding Gaussian is considered as a hard Gaussian and will grow. This strategy robustly identifies noticeable pixel rendering errors across views to mine hard Gaussians, thus reducing rendering errors in less observed regions.

Through our proposed strategies to grow and optimize hard Gaussians, our method helps to resolve blurring and needle-like artifacts, as shown in Figure~\ref{fig:teaser}(d). Extensive experiments on challenging benchmark datasets~\cite{barron2022mip,knapitsch2017tanks,hedman2018deep} demonstrate that our method achieves state-of-the-art rendering quality while maintaining real-time efficiency. In summary, our main contributions are as follows:
\begin{itemize}
    \item We analyze the deficiencies of Gaussian growing criterion--it smooths out significant positional gradients and rendering errors of Gaussians. This hinders hard Gaussians growing, causing artifacts in challenging areas. 
    \item We propose HGS, which mines hard Gaussians from multi-view significant positional gradients and rendering errors for growing and optimization. This alleviates cross-view rendering inconsistency and reduces rendering errors in less observed regions, significantly improving rendering performance.
    \item Our proposed HGS is general. In experiments, we show that it can boost both explicit Gaussian methods~\cite{kerbl20233d,yu2024mip,zhang2024pixel} and neural Gaussian methods~\cite{lu2024scaffold,chen2025hac} to achieve better NVS results. 
\end{itemize}

\section{Related Work}
\label{sec:related}

%-------------------------------------------------------------------------
\customparagraph{Novel View Synthesis} NVS aims to synthesize photo-realistic images from viewpoints different from the original input viewpoints. Early image-based rendering methods~\cite{chaurasia2013depth,hedman2018deep,riegler2020free,kopanas2021point} leverage 3D proxy geometry to generate novel views from source images. Over the past few years, NeRF~\cite{mildenhall2021nerf} has achieved impressive NVS results~\cite{barron2021mip,barron2022mip,fridovich2022plenoxels,verbin2022ref,yu2021plenoctrees,reiser2021kilonerf} and advanced other 3D tasks~\cite{poole2022dreamfusion,yariv2021volume,wang2021neus,fu2022geo,yi2024diffusion,sun2024global,xu2024fewshot}. NeRF utilizes a multi-layer perceptron (MLP) to model scenes and leverages ray marching to render images. Nevertheless, the expensive MLP evaluation caused by ray marching hinders the real-time performance. While subsequent methods adopt more efficient representations~\cite{muller2022instant,sun2022direct,chen2022tensorf,xu2022point,barron2023zip} to accelerate the rendering process, they still struggle with real-time requirements.

%-------------------------------------------------------------------------
\customparagraph{Gaussian Splatting} 3DGS~\cite{kerbl20233d} employs anisotropic Gaussians~\cite{zwicker2001ewa} to represent 3D scenes and realizes rendering by differentiable rasterization. It stands out for its real-time high-quality rendering results. Therefore, it has been rapidly extended to other 3D tasks, including surface reconstruction~\cite{yu2024gaussian,huang20242d}, physical simulation~\cite{xie2024physgaussian,zhang2025physdreamer} and 3D generation~\cite{tang2023dreamgaussian,yi2024mvgamba}. To improve 3DGS, some works tackle aliasing issues by introducing 3D smoothing and 2D Mip filters~\cite{yu2024mip}, multi-scale Gaussians~\cite{yan2024multi} and analytic integration~\cite{liang2024analytic}. Scaffold-GS~\cite{lu2024scaffold} builds anchor points to guide the distribution of local 3D Gaussians. Some methods~\cite{kerbl2024hierarchical,liu2025citygaussian} leverage Level-of-Detail for large-scale scene reconstruction. Although these methods enhance 3DGS, they cannot handle strong artifacts in challenging areas.

%-------------------------------------------------------------------------
\customparagraph{Blurring Artifacts} Since the 3D Gaussians initialized from sparse point clouds of SfM are too sparse to describe complete 3D scenes, 3DGS develops a growing strategy to densify them and improve rendering quality. However, there are still large empty areas cannot be populated due to the gradient collision~\cite{ye2024absgs}, resulting in blurring artifacts. To address this, some approaches~\cite{yu2024gaussian,ye2024absgs} accumulate the norms of individual pixel gradients or homodirectional positional gradients to better indicate regions with significant reconstruction errors. Pixel-GS~\cite{zhang2024pixel} considers the number of pixels covered by the Gaussian. By using these counts as weights to dynamically average the gradients from different viewpoints, it prompts the growth of large Gaussians. Mini-Splatting~\cite{fang2024mini} finds that the blurry artifacts are more directly related to the rendered index of Gaussians with the maximum contribution to each pixel. It then splits Gaussians with many rendered indices to reduce these artifacts. Spacetime-GS~\cite{li2024spacetime} improves rendering quality at distant sparsely covered areas by sampling new Gaussians with the guidance of training error and coarse depth. In addition, some methods~\cite{li2025geogaussian,zhu2025fsgs} resort to geometry priors to reduce artifacts. In contrast, our method mines hard Gaussians from multi-view significant positional gradients and rendering errors, and grow these Gaussians to resolve artifacts.

\section{Preliminaries}
\label{sec:preliminaries}

\customparagraph{3D Gaussian Splatting} 3DGS~\cite{kerbl20233d} represents a 3D scene with a set of anisotropic 3D Gaussians $\{\mathcal{G}_i | i=1,\cdots, N\}$ and renders an image using $\alpha$-blending with differentiable rasterization. Each Gaussian $\mathcal{G}_i$ is parameterized by a mean vector $\bm{\mu}_i \in \mathbb{R}^{3 \times 1}$, covariance matrix $\bm{\Sigma}_i \in \mathbb{R}^{3 \times 3}$, color $\mathbf{c}_i \in \mathbb{R}^{3 \times 1}$ and opacity $\alpha_i \in \mathbb{R}$ as:
\begin{equation}
    \mathcal{G}_i(\mathbf{x}) = \text{exp}(-\frac{1}{2}(\mathbf{x} - \bm{\mu}_i)^{T} \bm{\Sigma}_i^{-1} (\mathbf{x} - \bm{\mu}_i)),
\end{equation}
 where $\mathbf{x}$ is a spatial point within the 3D scene. To render an image from viewpoint $j$, the 3D Gaussian $\mathcal{G}_i$ is first splatted to the 2D image plane as a 2D Gaussian $\mathcal{G}_i^{2D}$, and then $\alpha$-blending is employed to compute the color of pixel $\bm{u}$ as:
\begin{equation}\label{eq:blending}
    \mathbf{c}(\bm{u}) = \sum_{i=1}^{N} w_i \mathbf{c}_i, \: w_i = \alpha_i \mathcal{G}_i^{2D}(\bm{u}) \prod_{k=1}^{i-1} (1 - \alpha_k \mathcal{G}_k^{2D}(\bm{u})).
\end{equation}
Through the differentiable rasterization, all the attributes in 3D Gaussians are optimized by the rendering loss. 3DGS designs an adaptive density control mechanism for Gaussian growing, adding new Gaussians where significant Gaussians are found.

\customparagraph{Scaffold-GS} As a neural Gaussian method, Scaffold-GS~\cite{lu2024scaffold} introduces anchor points to distribute local 3D Gaussians. Each anchor is associated with a location $\mathbf{x} \in \mathbb{R}^{3}$ and anchor attributes $\mathcal{A}=\{\bm{f} \in \mathbb{R}^{32}, \bm{l} \in \mathbb{R}^{3}, \bm{o} \in \mathbb{R}^{K \times 3}\}$,  where each component represents anchor context feature, scaling factor and offsets, respectively. The positions of $K$ neural Gaussians are calculated as:
\begin{equation}\label{eq:scaffold}
    \{\bm{\mu}_1, \cdots, \bm{\mu}_K \} = \bm{x} + \bm{o} \cdot \bm{l}.
\end{equation}
The properties of neural Gaussians are decoded from the anchor attributes, and then used for $\alpha$-blending to render images. During the Gaussian growing stage, Scaffold-GS constructs voxel grids to find significant neural Gaussians and grow their corresponding anchor points.

\begin{figure*}[htp]
    \centering
    \includegraphics[width=\linewidth]{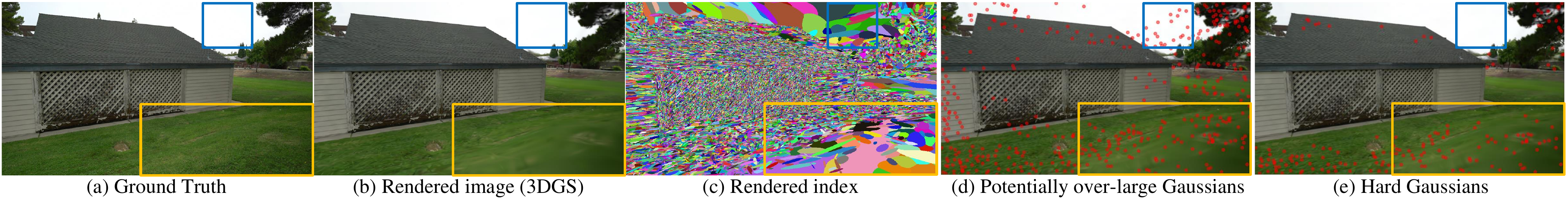}
    \caption{\textbf{Visual analysis of rendering error guided HGS.} 
    (a) Ground truth image. (b) Rendered image by 3DGS. (c) Rendered index of Gaussians with the maximum contribution to each pixel. (d) Projection points of potentially over-large Gaussians. (e) Projection points of hard Gaussians. Potentially over-large Gaussians may indicate false positives for Gaussian growing while hard Gaussians alleviate this (the blue boxes in (d) and (e)).}
    \label{fig:mv-hem-re}
\end{figure*}

\section{Deficiencies in Gaussian Growing}
\label{sec:deficiencies}

Both explicit Gaussian methods and neural Gaussian methods leverage sparse point clouds from SfM to initialize Gaussians or anchor points, which cannot completely model a 3D scene. To make explicit/neural Gaussians better represent the scene, Gaussian growing is applied to densify existing Gaussians gradually. Since the two kinds of methods both leverage significant Gaussians for Gaussian growing, we detail how to find significant Gaussians below.
% For simplify, we detail   

Specifically, one Gaussian will be deemed as significant when its average view-space positional gradient over $M$ training iterations (one growth interval) satisfies:
\begin{equation}\label{eq:grow}
    \frac{\sum_j||\bm{g}_{i,j}||}{n} \geq \tau_\text{grad}, 
\end{equation}
where $\bm{g}_{i,j}$ denotes the view-space positional gradient of Gaussian $\mathcal{G}_i$ under viewpoint $j$, $n$ is the total number of viewpoints that Gaussian $\mathcal{G}_i$ participates in the calculation during the $M$ iterations, and $\tau_\text{grad}$ is the gradient threshold. Due to the smoothing effect of the average operation in Eq.~\eqref{eq:grow}, some individual larger view-space positional gradients cannot drive Gaussians to grow, as shown in Figure~\ref{fig:motivation}. This makes Gaussian growing focus on most easy-to-learn viewpoints and ignore some hard-to-learn viewpoints, leading to cross-view rendering inconsistency.

Moreover, the view-space positional gradient is computed by:
\begin{equation}\label{eq:pg}
    \bm{g}_{i,j} = \frac{\partial L_j}{\partial \bm{\mu}_{i,j}}, 
    \quad L_j = \frac{\sum_{\bm{u} \in \bm{P}} e_{\bm{u}}}{||\bm{P}||}, 
\end{equation}
where $L_i$ denotes the rendering loss under viewpoint $j$, $\bm{\mu}_{i,j}$ is the pixel-space projection point of Gaussian $\mathcal{G}_i$ under viewpoint $j$, $\bm{P}$ is the total pixel coordinates, and $e_{\bm{u}}$ is the rendering error of pixel $\bm{u}$. Since $L_j$ indicates the rendering quality of the entire image instead of local image regions, the resulting average view-space positional gradient fails to reflect lager local rendering errors. This prevents Gaussians from growing in regions with larger local rendering errors. 

One straightforward way to alleviate the above problems is to lower the gradient threshold $\tau_\text{grad}$ to encourage more Gaussians to grow. However, as pointed out in \cite{zhang2024pixel,ye2024absgs}, 
lowering $\tau_\text{grad}$ will prioritize densifying the well-reconstructed areas. This will introduce \emph{redundant Gaussians} in these areas and cannot effectively reduce artifacts (See supplementary for more analysis).  
 
%-------------------------------------------------------------------------

\section{Hard Gaussians Splatting--HGS}
\label{sec:method}
 
To address the problems analyzed above, we propose Hard Gaussian Splatting, dubbed HGS, to uncover hard Gaussians from multi-view significant positional gradients and rendering errors. In this way, our method can grow and optimize these hard Gaussians to recover more complete 3D scenes, thereby improving rendering quality.

%-------------------------------------------------------------------------
\subsection{Positional Gradient Driven HGS}
\label{sec:MH-PG}

The view-space positional gradient reflects the overall image reconstruction quality under a certain viewpoint. The original Gaussian growing criterion averages the $n$ view-space positional gradients over $M$ iterations to find growing candidates. Although the average operation can reduce the impact of noise, it smooths some individual larger positional gradients. 

In fact, the larger view-space positional gradients also indicate that their corresponding Gaussians need to populate empty areas. However, they are ignored by the original growing criterion. Therefore, we present Positional Gradient driven HGS (PGHGS), which uncovers hard Gaussians from multi-view significant positional gradients. To reliably capture significant positional gradients, we first sort the $n$ positional gradients for Gaussian $\mathcal{G}_i$ as:
\begin{equation}
    \{\bm{g}'_{i,1}, \cdots, \bm{g}'_{i,n}\} = \text{sort}_{\downarrow} (\bm{g}_{i,1}, \cdots, \bm{g}_{i,n}),
\end{equation}
where $\text{sort}_{\downarrow}$ denotes descending sort. Then, we mine hard Gaussians if the $\textit{k-th}$ largest positional gradient satisfies:
\begin{equation}\label{eq:mpg}
    ||\bm{g}'_{i,k}|| \geq \lambda\tau_\text{grad},
\end{equation}
where $\lambda$ is a constant that controls the extent of excavation. By growing these hard Gaussians, the larger reconstruction errors under some individual viewpoints can be reduced. This facilitates the rendering consistency across views.

%-------------------------------------------------------------------------
\subsection{Rendering Error Guided HGS}
\label{sec:MH-RE}

By delving into the calculation of view-space positional gradient (Eq.~\eqref{eq:pg}), we know that it is the gradient of the average rendering error with respect to the view-space projection point. Therefore, the view-space positional gradient cannot effectively reflect the reconstruction errors of some local image regions, especially when these regions are only observed by few viewpoints. To resolve this, we introduce Rendering Error guided HGS (REHGS) to link hard Gaussians with local rendering errors. However, as pixel rendering errors entangle contributions from multiple Gaussians (Eq.~\eqref{eq:blending}), it is challenging to establish this link. 

To this end, we leverage the Gaussians with the maximum contribution to correlate pixel rendering errors. 
Specifically, for pixel $\bm{u}$, its rendered index is defined as $\text{idx}(\bm{u}) = \mathop{\arg\max}\limits_{i} w_i$. 
This considers the maximum Gaussian contribution in $\alpha$-blending, thus reflecting the relationship between pixel rendering errors and Gaussians to some extent. On this basis, Gaussian $\mathcal{G}_i$ has the maximum rendering contribution on these pixels $\bm{S}_i = \{\bm{u} | \text{idx}(\bm{u})=i, \bm{u} \in \bm{P}\}$. To detect noticeable local rendering errors rather than outlier pixels, we first identify potentially over-large Gaussians based on the size of $\bm{S}_i$. One Gaussian will be considered potentially over-large and prone to blurring artifacts if it meets the following condition~\cite{fang2024mini}:
\begin{equation}\label{eq:large}
    ||\bm{S}_i|| > \tau_\text{large} ||\bm{P}||, 
\end{equation}
where $\tau_\text{large}$ is a threshold to control the extent of potentially large areas. This condition can efficiently find Gaussians which may lead to burring artifacts (``over-reconstruction''), as shown in the yellow boxes of Figures~\ref{fig:mv-hem-re}(c) and (d).

In fact, it is possible to describe low-textured areas and repetitive textures with potentially over-large Gaussians, as shown in the blue box of Figure~\ref{fig:mv-hem-re}(d). Therefore, using only Eq.~\eqref{eq:large} to determine the over-large Gaussians will result in false positives. To locate the Gaussians that cause blurring artifacts more accurately, we leverage rendering errors to mine hard Gaussians. For one potentially over-large Gaussian $\mathcal{G}_i$, we project it into the current training viewpoint $j$ to obtain the corresponding pixel $\bm{u}_{i,j}$. Then, this Gaussian will be considered as a hard Gaussian if it satisfies:
\begin{equation}\label{eq:re}
    \text{SSIM} (\bm{I}_j, \hat{\bm{I}}_j)(\bm{u}_{i,j}) < \tau_\text{SSIM},
\end{equation}
where $\bm{I}_j$ and $\hat{\bm{I}}_j$ denote the ground truth and rendered images under viewpoint $j$, respectively. SSIM means Structural Similarity Index Measure~\cite{wang2004image}, which measures the similarity between two images. $\tau_\text{SSIM}$ is a threshold to judge rendering quality. Note that, SSIM is chosen because it can perform local analyssis on an image, making it suitable for detecting structure changes in the image.

With the guidance of potentially over-large Gaussians and pixel rendering errors, our method can locate hard Gaussians in truly blurring areas, as shown in Figure~\ref{fig:mv-hem-re}(e).  
Furthermore, such Gaussians should be seen by at least two views during the $M$ iterations. This avoids unstable growth caused by hard Gaussians determined by only a single view. 

\subsection{Efficient HGS}

In practice, as more hard Gaussians are grown and optimized, the efficiency of Gaussian splatting will be sacrificed. To better balance rendering quality and efficiency, we introduce an efficient HGS method, called Effi-HGS. Concretely, we first count the percentage of growing Gaussians selected by the original growing criterion (Eq.~\eqref{eq:grow}) in each growth interval, denoted as $Q$.  Then, during the PGHGS, we choose the top $Q$ Gaussians as hard ones for growing based on $\{\bm{g}'_{i,k} | i=1, \cdots, N\}$. In this way, Effi-HGS can adaptively control the number of hard Gaussians for efficiency while improving rendering quality. 
\section{Experiments}
\label{sec:experiment}

\subsection{Experimental Setup}

\customparagraph{Datasets and Metrics} We comprehensively evaluate our methods across 17 scenes from publicly available datasets, including nine scenes from Mip-NeRF360~\cite{barron2022mip}, six scenes from Tanks\&Temples~\cite{knapitsch2017tanks} and two scenes from Deep Blending~\cite{hedman2018deep}. The first two datasets contain both bounded indoor scenes and unbounded outdoor scenes. The last one is with two bounded indoor scenes. We evaluate the quality of synthesized novel views through several quantitative metrics, including Peak Signal-to-Noise Ratio (PSNR), Structural Similarity Index Measure (SSIM)~\cite{wang2004image} and Learned Perceptual Image Patch Similarity (LPIPS)~\cite{zhang2018unreasonable}.   

\customparagraph{Baselines and Implementations} We compare our methods against state-of-the-art NeRF and Gaussian splatting methods. The NeRF methods include Plenoxels~\cite{fridovich2022plenoxels}, INGP~\cite{muller2022instant} and Mip-NeRF360~\cite{barron2022mip}, and their evaluation results are from \cite{kerbl20233d}. For the Gaussian splatting methods, explicit methods, including 3DGS~\cite{kerbl20233d}, Mip-Splatting\footnote{The Mip-Splatting we compared is the version that has been combined with the densification in GOF~\cite{yu2024gaussian}.}~\cite{yu2024mip} and Pixel-GS~\cite{zhang2024pixel}, and the neural method, Scaffold-GS~\cite{lu2024scaffold}, are compared. We retrain these Gaussian splatting methods for 30K iterations on the same platform. We build our methods upon the Scaffold-GS repository~\cite{lu2024scaffold}. We maintain the same threshold $\tau_\text{grad}$ and growing interval $M$ for Gaussian growing as in the Scaffold-GS. We set $\{k, \lambda, \tau_\text{large}, \tau_\text{SSIM}\} = \{3, 1.0, 2\text{e-4}, 0.7\}$ across all scenes. All experiments are conducted on one NVIDIA RTX 3090 GPU with 24GB memory.

\subsection{Performance Evaluation}
\label{sec:evaluation}

\begin{table*}[t]
    \centering
    % \small
    \resizebox{\linewidth}{!}{
\begin{tabular}{l|ccc|ccc|ccc}
\toprule
Dataset & \multicolumn{3}{c|}{Mip-NeRF360} & \multicolumn{3}{c|}{Tanks\&Temples} & \multicolumn{3}{c}{Deep Blending} \\
Method$\,|\,$Metric & PSNR \(\uparrow\) & SSIM \(\uparrow\) & LPIPS \(\downarrow\) & PSNR \(\uparrow\) & SSIM \(\uparrow\) & LPIPS \(\downarrow\) & PSNR \(\uparrow\) & SSIM \(\uparrow\) & LPIPS \(\downarrow\) \\
\midrule
Plenoxels~\cite{fridovich2022plenoxels} & 23.08 & 0.626 & 0.463 & - & - & - & 23.06 & 0.795 & 0.510 \\ 
INGP~\cite{muller2022instant} & 25.59 & 0.699 & 0.331 & - & - & - & 23.62 & 0.797 & 0.423 \\
Mip-NeRF360~\cite{barron2022mip} & 27.69 & 0.792 & 0.237 & - & - & - & 29.40 & 0.901 & 0.245 \\
3DGS~\cite{kerbl20233d} & 27.71 & 0.825 & 0.202 & 24.90 & 0.850 & 0.178 & 29.46 & 0.899 & 0.247 \\
% mini-Splatting~\cite{fang2024mini} & 27.59 & 0.839 & 0.164 & 24.20 & 0.846 & 0.162 & 29.88 & 0.906 & 0.211 \\
Pixel-GS~\cite{zhang2024pixel} & 27.87 & \cellcolor{tabthird}0.835 & \cellcolor{tabthird}0.176 & 25.18 & 0.860 & \cellcolor{tabthird}0.157 & 28.88 & 0.892 & 0.251 \\
Mip-Splatting~\cite{yu2024mip} & 27.92 & \cellcolor{tabfirst}0.838 & \cellcolor{tabsecond}0.175 & 25.02 & 0.857 & 0.166 & 29.25 & \cellcolor{tabthird}0.903 & \cellcolor{tabthird}0.240 \\
Scaffold-GS~\cite{lu2024scaffold} & \cellcolor{tabthird}27.98 & 0.825 & 0.208 & \cellcolor{tabthird}25.74 & \cellcolor{tabthird}0.861 & 0.170 & \cellcolor{tabthird}30.23 & \cellcolor{tabsecond}0.907 & 0.245 \\
\midrule
% \textbf{MH-GS (Ours)} & 28.15 & 0.830 & 0.176 & 26.13 & 0.873 & 0.140 & 30.35 & 0.908 & 0.230 \\
\textbf{Effi-HGS (Ours)} & \cellcolor{tabsecond}28.14 & 0.831 & 0.179 & \cellcolor{tabsecond}26.02 & \cellcolor{tabsecond}0.870 & \cellcolor{tabsecond}0.148 & \cellcolor{tabsecond}30.33 & \cellcolor{tabfirst}0.908 & \cellcolor{tabsecond}0.238 \\
\textbf{HGS (Ours)} & \cellcolor{tabfirst}28.30 & \cellcolor{tabsecond}0.837 & \cellcolor{tabfirst}0.166 & \cellcolor{tabfirst}26.36 & \cellcolor{tabfirst}0.880 & \cellcolor{tabfirst}0.126  & \cellcolor{tabfirst}30.40 & \cellcolor{tabfirst}0.908 & \cellcolor{tabfirst}0.225 \\
\bottomrule
\end{tabular}}
    \caption{\textbf{Quantitative comparisons on three datasets~\cite{barron2022mip,knapitsch2017tanks,hedman2018deep}.} The best, second-best, and third-best entries are marked in red, orange, and yellow, respectively.}
    \label{tab:quan_comp}
\end{table*}

\begin{figure*}
    \centering
    \includegraphics[width=\linewidth]{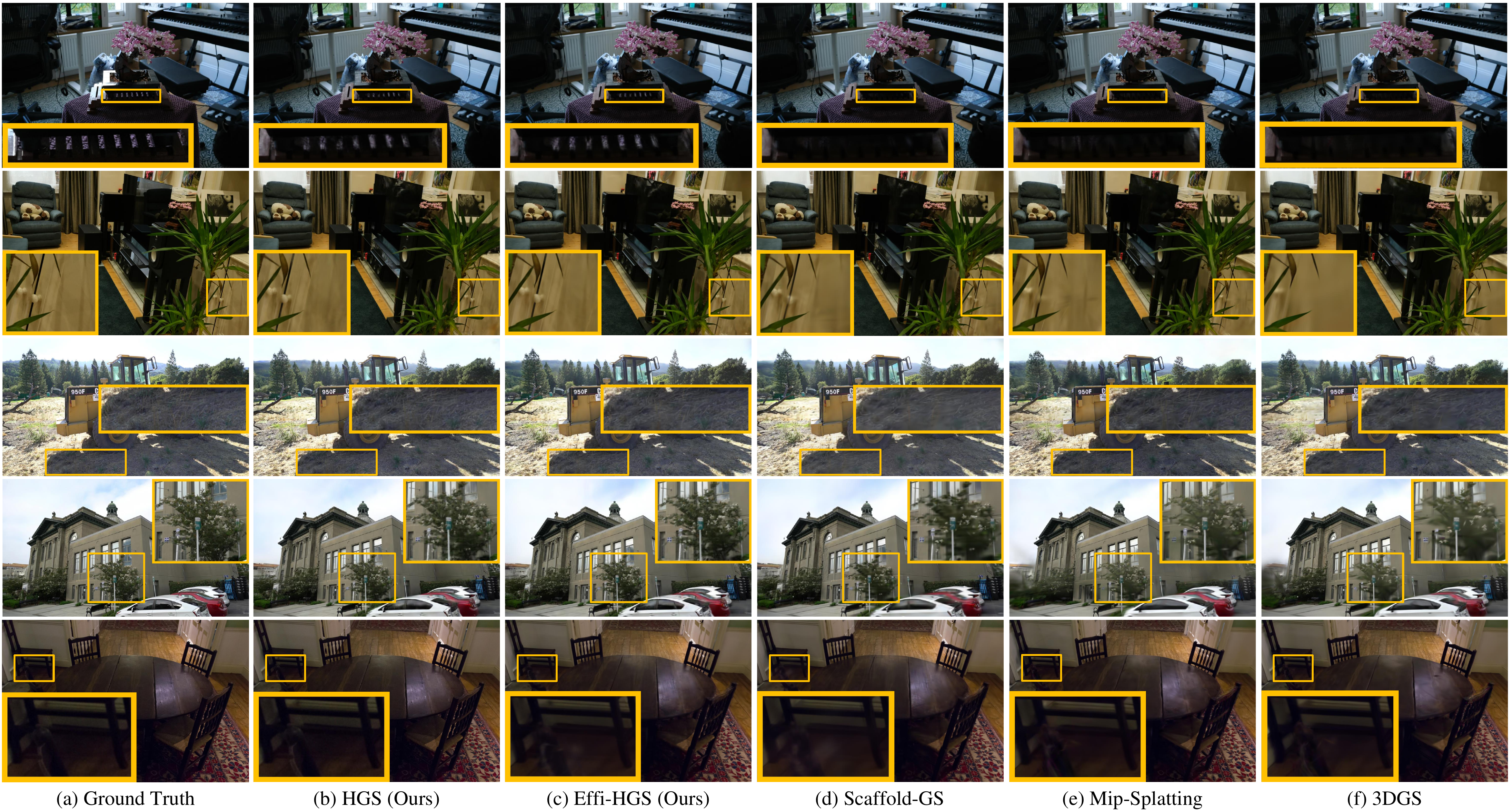}
    \caption{\textbf{Qualitative comparisons on three datasets~\cite{barron2022mip,knapitsch2017tanks,hedman2018deep}.} Yellow boxes show challenging areas. Our methods can reconstruct these areas well while other methods fail.}
    \label{fig:vis_comp}
\end{figure*}

\customparagraph{Rendering Quality} The quantitative results on three datasets~\cite{barron2022mip,knapitsch2017tanks,hedman2018deep} are presented in Table~\ref{tab:quan_comp}. It can be noticed that our proposed method, HGS, achieves almost the best rendering results in all metrics on these three datasets, expect for the second best SSIM on Mip-NeRF360. Notably, HGS outperforms the state-of-the-arts by a large margin on Tanks\&Temples, which contains large-scale complex scenes. Moreover, our efficient version, Effi-HGS, achieves comparable rendering results with the state-of-the-arts on Mip-NeRF360, and surpasses them on Tanks\&Temples and Deep Blending.

The qualitative results are shown in Figure~\ref{fig:vis_comp}. We observe that both our proposed methods significantly reduce blurring and needle-like artifacts. In particular, it is challenging for the previous methods to reconstruct some occluded areas, such as the the top row in Figure~\ref{fig:vis_comp}. In contrast, our methods can recover these challenging areas. This can be attributed to that our approaches can mine hard Gaussians from significant positional gradients and rendering errors, and grow these Gaussians to optimize fine-scale and less observed areas. 

\customparagraph{Efficiency Comparisons} We further compare efficiency with Gaussian splatting methods in terms of training time, rendering speed and storage size. The results are reported in Table~\ref{tab:efficiency}. As can be seen, for explicit Gaussian methods, improved rendering quality requires growing more Gaussians (larger storage size), which increases training time and reduces rendering speed. The neural Gaussian method, Scaffold-GS, can enhance both rendering quality and efficiency compared to Mip-Splatting and Pixel-GS. Similarly, since our methods will grow many hard Gaussians to improve rendering quality, this will also increase training time and reduce rendering speed. Overall, the efficiency of our methods is acceptable considering the superior rendering quality. Moreover, our efficient one achieves competitive efficiency while showing much better rendering quality than previous methods.

\begin{table}[t]
    \centering
    \resizebox{\linewidth}{!}{
\begin{tabular}{l|ccc|ccc}
\toprule
& {PSNR $\uparrow$} & {SSIM $\uparrow$} & {LPIPS $\downarrow$} & Train & FPS & Memory \\
\midrule
3DGS~\cite{kerbl20233d} & 24.90 & 0.850 & 0.178 & \bf 15.2m & \bf 159 & 0.36GB \\
Pixel-GS~\cite{zhang2024pixel} & 25.18 & 0.860 & 0.157 & 26.3m & 88 & 0.80GB \\
Mip-Splatting~\cite{yu2024mip} & 25.02 & 0.857 & 0.166 & 20.0m & 126 & 0.45GB \\
Scaffold-GS~\cite{lu2024scaffold} & 25.74 & 0.861 & 0.170 & 20.4m & 125 & \bf 0.18GB \\
\midrule
\textbf{Effi-HGS (Ours)} & 26.02 & 0.870 & 0.148 & 28.6m & 87 & 0.30GB \\
\textbf{HGS (Ours)} & \bf 26.36 & \bf 0.880 & \bf 0.126 & 66.2m & 42 & 0.73GB \\
\bottomrule
\end{tabular}}
    \caption{\textbf{Efficiency comparisons on Tanks\&Temples~\cite{knapitsch2017tanks}.}}
    \label{tab:efficiency}
\end{table}

\subsection{Generalization Performance}

\begin{table}[t]
    \centering
    \resizebox{\linewidth}{!}
    {
\begin{tabular}{l|ccc}
\toprule
Method & {PSNR $\uparrow$} & {SSIM $\uparrow$} & {LPIPS $\downarrow$} \\
\midrule
3DGS~\cite{kerbl20233d} & 24.90 & 0.850 & 0.178 \\
Pixel-GS*~\cite{zhang2024pixel} & 25.26 & 0.861 & 0.158 \\
Mip-Splatting*~\cite{yu2024mip} & 25.17 & 0.861 & 0.163 \\
HAC~\cite{chen2025hac} & 25.77 & 0.859 & 0.173 \\
\midrule\midrule
\multirow{2}{*}{3DGS + HGS} & 25.04 & 0.856 & 0.163 \\
& 0.14 $\uparrow$ & 0.006 $\uparrow$ & 0.015 $\downarrow$ \\
\midrule
\multirow{2}{*}{Pixel-GS* + HGS} & 25.28 & 0.864 & \bf 0.147 \\
& 0.02 $\uparrow$ & 0.003 $\uparrow$ & 0.011 $\downarrow$ \\
\midrule
\multirow{2}{*}{Mip-Splatting* + HGS} & 25.26 & 0.862 & 0.155 \\
& 0.09 $\uparrow$ & 0.001 $\uparrow$ & 0.008 $\downarrow$ \\
\midrule
\multirow{2}{*}{HAC + HGS} & \bf 25.96 & \bf 0.869 & 0.150 \\
& 0.19 $\uparrow$ & 0.010 $\uparrow$ & 0.023 $\downarrow$ \\
\bottomrule\bottomrule
\end{tabular}}
    \caption{\textbf{Generalization performance of HGS on Tanks\&Temples~\cite{knapitsch2017tanks}.} Pixel-GS* and Mip-Splatting* mean they are combined with opacity correction~\cite{bulo2024revising}.}
    \label{tab:general}
\end{table}

To verify the generalization of our proposed approach, several state-of-the art Gaussian splatting methods are integrated with our proposed HGS to evaluate NVS on Tanks\&Temples. The considered methods include 3DGS, Pixel-GS, Mip-Splatting and HAC~\cite{chen2025hac}. Among them, the first three methods are explicit Gaussian methods. HAC is a compact neural Gaussian method that learns structured compact hash grids for anchor attributes modeling. Considering training efficiency, we test the generalization performance of our efficient HGS. Results are reported in Table~\ref{tab:general}.  

It can be seen that our proposed HGS boosts all rendering metrics for both explicit and neural Gaussian methods. Remarkably, HGS dramatically improves the LPIPS metric. Moreover, the performance improvement on neural Gaussian methods are more pronounced than on explicit Gaussian methods, as neural Gaussian methods will grow $K$ Gaussians for each added anchor point (Eq.~\eqref{eq:scaffold}). With our HGS, neural Gaussian methods greatly promote the Gaussian growing in poorly reconstructed regions, resulting in a more substantial performance improvement. 

Note that, for Pixel-GS and Mip-Splatting that have grown more Gaussians than 3DGS with their modified gradients, in order to improve their rendering quality with our HGS, we find that it is necessary to combine them with the opacity correction~\cite{bulo2024revising} to remove the bias in the cloning operation first. This is because the bias will hinder the Gaussian optimization more obviously when a large number of Gaussian points are cloned simultaneously. Based on the improvement brought by the opacity correction, our HGS can further boost the rendering performance. The results suggest that our HGS is general for both explicit and neural Gaussian methods.    

\subsection{Analysis}
\label{sec:analysis}

In this section, we perform ablation studies and analysis experiments on both Tanks\&Temples and Deep Blending to analyze the effectiveness of our proposed designs in Sec.~\ref{sec:method}.

\customparagraph{Ablation Study} We adopt Scaffold-GS as our baseline. Our different designs are progressively added to the baseline to investigate their efficacy. Results are reported in Table~\ref{tab:ablation} (See supplementary for ablation study on Deep Blending). By adding PGHGS to the baseline, the rendering quality of Model-A has been improved a lot in all metrics (PSNR: 0.58 $\uparrow$, SSIM: 0.018 $\uparrow$, LPIPS: 0.041 $\downarrow$). As for REHGS, it boosts the baseline obviously and further enhances Model-A. This demonstrates that both our proposed designs can mine hard Gaussians to improve rendering quality. In addition, compared to PGHGS, REHGS tends to find noticeable rendering errors, it cannot detect some inconspicuous artifacts, which can be probed better by PGHGS. Thus, the improvement brought by PGHGS is more obvious. 

\begin{figure*}[t]
    \centering
    \includegraphics[width=\linewidth]{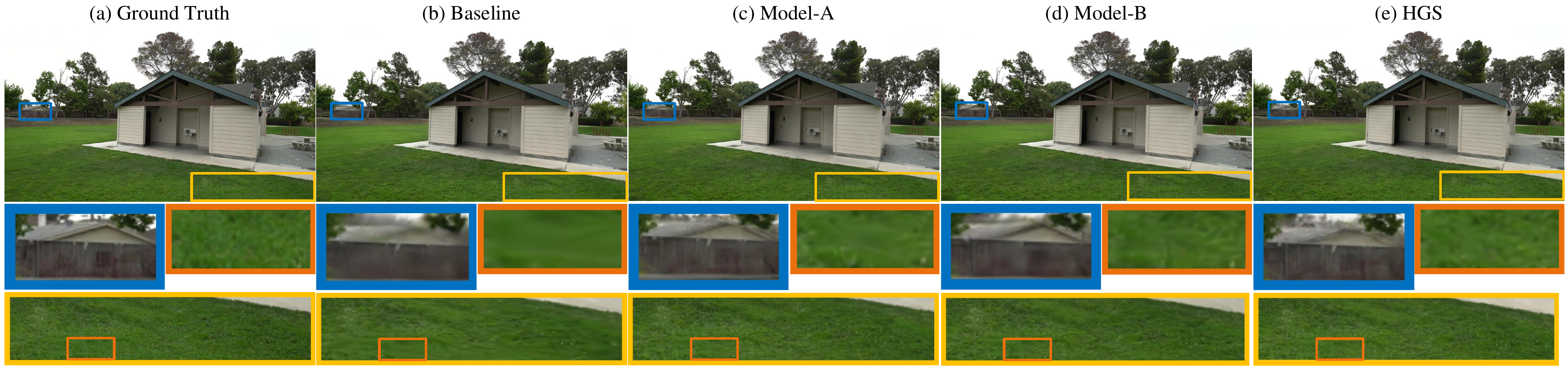}
    \caption{\textbf{Qualitative results of ablation study.}}
    \label{fig:ablation}
\end{figure*}

Figure~\ref{fig:ablation} shows how the proposed designs
improve the rendering quality. Both Model-A and Model-B can reconstruct details better than Baseline, such as the distant house and front grass in the blue and yellow boxes of Figure~\ref{fig:ablation}. Moreover, our full model can further reduce blurring artifacts in Model-A, such as the orange boxes in Figures~\ref{fig:ablation}(c) and (e). These results show that our method effectively improves rendering performance.

\begin{table}[t]
    \centering
    \resizebox{\linewidth}{!}{

\begin{tabular}{l|ccc|ccc}
\toprule
& OG & PGHGS & REHGS & {PSNR $\uparrow$} & {SSIM $\uparrow$} & {LPIPS $\downarrow$} \\
\midrule
Baseline & \checkmark & & & 25.74 & 0.861 & 0.170 \\
\midrule
Model-A & \checkmark & \checkmark & & 26.32 & 0.879 & 0.129 \\
Model-B &  \checkmark & & \checkmark & 25.99 & 0.867 & 0.155 \\
HGS & \checkmark & \checkmark & \checkmark & \bf 26.36 & \bf 0.880 & \bf 0.126  \\
\bottomrule
\end{tabular}}
    \caption{\textbf{Ablation study on Tanks\&Temples~\cite{knapitsch2017tanks}.} OG: Original Growing criterion. PGHGS: Positional Gradient driven HGS. REHGS: Rendering Error guided HGS.}
    \label{tab:ablation}
\end{table}

\customparagraph{Effect of $k$} $k$ represents the number of significant positional gradients in the growth interval in Eq.~\eqref{eq:mpg}. We vary this value to study its effect and report the results in Table~\ref{tab:analysis}. We observe that: 1) Reducing $k$ can further improve LPIPS but degrade PSNR. This is because reducing $k$ relaxes the criterion for significant positional gradients, leading to more growing Gaussians. This may introduce noise to impair PSNR. On the other hand, more Gaussians are helpful to alleviate blurring artifacts, and LPIPS is evaluated through deep image features which have anti-noise ability. Therefore, reducing $k$ can improve LPIPS. 2) Increasing $k$ degrades almost all rendering metrics, suggesting that PGHGS cannot mine hard Gaussians sufficiently when $k$ is too large. 

\begin{table}[t]
    \centering
    \resizebox{\linewidth}{!}{

\begin{tabular}{l|c|ccc|ccc}
\toprule
\multirow{2}{*}{Module} & \multirow{2}{*}{Setting} & \multicolumn{3}{c|}{Tanks\&Temples} & \multicolumn{3}{c}{Deep Blending} \\
& & {PSNR $\uparrow$} & {SSIM $\uparrow$} & {LPIPS $\downarrow$} & {PSNR $\uparrow$} & {SSIM $\uparrow$} & {LPIPS $\downarrow$} \\
\midrule
\multirow{4}{*}{PGHGS} & $k$=1 & - & - & - & 30.31 & 0.906 & \bf 0.211 \\
& $k$=2 & 26.34 & \bf 0.880 & \bf 0.123 & 30.39 & \bf 0.908 & 0.219 \\
& $k$=3 & \bf 26.36 & \bf 0.880 & 0.126 & \bf 30.40 & \bf 0.908 & 0.225 \\
& $k$=4 & 26.26 & 0.878 & 0.131 & 30.19 & \bf 0.908 & 0.230 \\
\midrule
\multirow{8}{*}{REHGS} & POG & 26.26 & 0.879 & \bf 0.121 & 30.17 & 0.896 & \bf 0.194 \\
& HG & \bf 26.36 & \bf 0.880 & 0.126 & \bf 30.40 & \bf 0.908 & 0.225 \\
\cmidrule(r){2-8}
& $\tau_\text{SSIM}$=0.6 & 26.34 & 0.880 & 0.127 & 30.29 & 0.908 & 0.225 \\
& $\tau_\text{SSIM}$=0.7 & 26.36 & 0.880 & \bf 0.126 & \bf 30.40 & 0.908 & 0.225 \\
& $\tau_\text{SSIM}$=0.8 & \bf 26.39 & 0.880 & \bf 0.126 & 30.37 & 0.908 & \bf 0.224 \\
\cmidrule(r){2-8}
& $\tau_\text{large}$=1e-4 & \bf 26.38 & \bf 0.880 & \bf 0.125 & 30.35 & 0.908 & 0.225 \\
& $\tau_\text{large}$=2e-4 & 26.36 & \bf 0.880 & 0.126 & \bf 30.40 & 0.908 & 0.225 \\
& $\tau_\text{large}$=3e-4 & 26.36 & 0.879 & 0.127 & 30.39 & 0.908 & 0.225 \\
\bottomrule
\end{tabular}}
    \caption{\textbf{More module analysis on both Tanks\&Temples~\cite{knapitsch2017tanks} and Deep Blending~\cite{hedman2018deep}.} PGHGS: Positional Gradient driven HGS. REHGS: Rendering Error guided HGS. POG: Potentially Over-large Gaussians. HG: Hard Gaussians. - indicates no results due to out of memory error.}
    \label{tab:analysis}
\end{table}

\begin{figure}
    \centering
    \includegraphics[width=\linewidth]{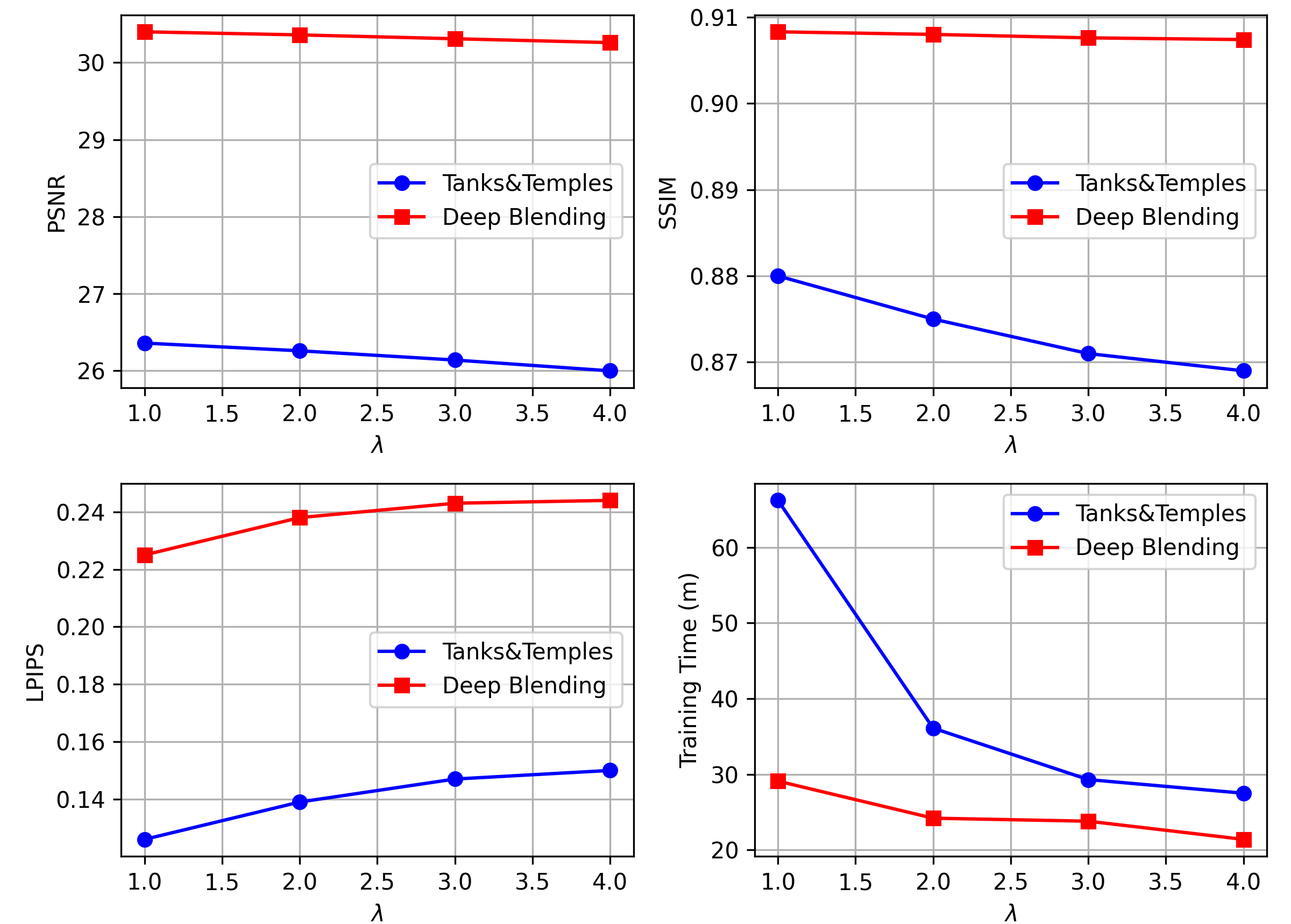}
    \caption{\textbf{Effect of $\lambda$.}}
    \label{fig:lambda}
\end{figure}

\customparagraph{Effect of $\lambda$} 
% $\lambda$ controls the extent of  excavation in Eq~\eqref{eq:mpg}. 
To investigate the impact of $\lambda$ in Eq~\eqref{eq:mpg}, we set $\lambda$ to different values and show the results in Figure~\ref{fig:lambda}. As the $\lambda$ increases, the rendering quality (PSNR, SSIM and LPIPS) degrades while the training efficiency improves. This suggests that lowering $\lambda$ allows for mining as many hard Gaussians as possible to improve rendering quality. Therefore, the training efficiency is reduced when a large number of Gaussians are splatted and optimized.

\customparagraph{Potentially Over-large Gaussians or Hard Gaussians} In Sec.~\ref{sec:MH-RE}, we locate hard Gaussians from potentially over-large Gaussians. Here, we compare the impact of potentially over-large Gaussians (Eq.~\eqref{eq:large}) and hard Gaussians (Eq.~\eqref{eq:re}) on rendering performance in Table~\ref{tab:ablation}. Potentially over-large Gaussians improve LPIPS slightly but degrade PSNR a lot. As demonstrated in Sec.~\ref{sec:MH-RE}, potentially over-large Gaussians will overgrow the already learned areas. This makes the already learned areas susceptible to redundant Gaussians, resulting in the PSNR degradation.  

\customparagraph{Effect of $\tau_\text{SSIM}$} We evaluate the NVS performance with different $\tau_\text{SSIM}$ values in Table~\ref{tab:analysis}. Lowering $\tau_\text{SSIM}$ prevents REHGS from fully identifying noticeable rendering errors, degrading NVS performance. The performance is nearly optimal with our set value, and converges at larger $\tau_\text{SSIM}$.    

\customparagraph{Effect of $\tau_\text{large}$} We further test the NVS performance with different $\tau_\text{large}$ values in Table~\ref{tab:analysis}. This parameter is almost insensitive to these varying values. Moreover, our selected value almost achieves the best NVS performance on both Tanks\&Temples and Deep Blending.  
\section{Conclusion}
\label{sec:conclusion}

In this work, we presented HGS, a hard Gaussians growing framework for Gaussian Splatting, which leverages multi-view significant positional gradients and rendering errors to uncover hard Gaussians. By growing and optimizing these hard Gaussians, our methods can effectively resolve blurring and needle-like artifacts in challenging areas. Our experimental results demonstrate that HGS achieves superior rendering quality while maintaining real-time rendering speed. Moreover, we verify that our HGS can boost both explicit and neural Gaussian methods, showing its potential generalization ability to other Gaussian splatting related tasks.   
{
    \small
    \bibliographystyle{ieeenat_fullname}
    \bibliography{main}
}

\clearpage
\setcounter{page}{1}
\maketitlesupplementary

In this \textbf{supplementary document}, we first provide more analysis for the original Gaussian growing in Section~\ref{sec:more_analysis}. Then, we report implementation details, additional ablation study and experimental results in Section~\ref{sec:implementation}, Section~\ref{sec:ablation_db} and Section~\ref{sec:additional}, respectively. Next, we show the anti-aliasing ability of Mip-Splatting~\cite{yu2024mip} when combined with our method in Section~\ref{sec:anti-aliasing}. Finally, we present discussions and limitations in Section~\ref{sec:discussion}.

\section{More Analysis of Gaussian Growing}
\label{sec:more_analysis}

Since the strong artifacts in some areas can be attributed to insufficient Gaussians in these areas, one possible way to encourage Gaussians growing in these areas is to lower the gradient threshold $\tau_\text{grad}$ in Eq.~\eqref{eq:grow}. However, previous studies~\cite{zhang2024pixel,ye2024absgs} have pointed out that lowering the threshold for 3DGS~\cite{kerbl20233d} will preferentially grow the Gaussians in well-reconstructed areas. This cannot effectively alleviate the artifacts. Here, we further study the impact of lowering $\tau_\text{grad}$ for the neural Gaussian method, Scaffold-GS~\cite{lu2024scaffold}. 

\begin{figure*}[h]
    \centering
    \includegraphics[width=\linewidth]{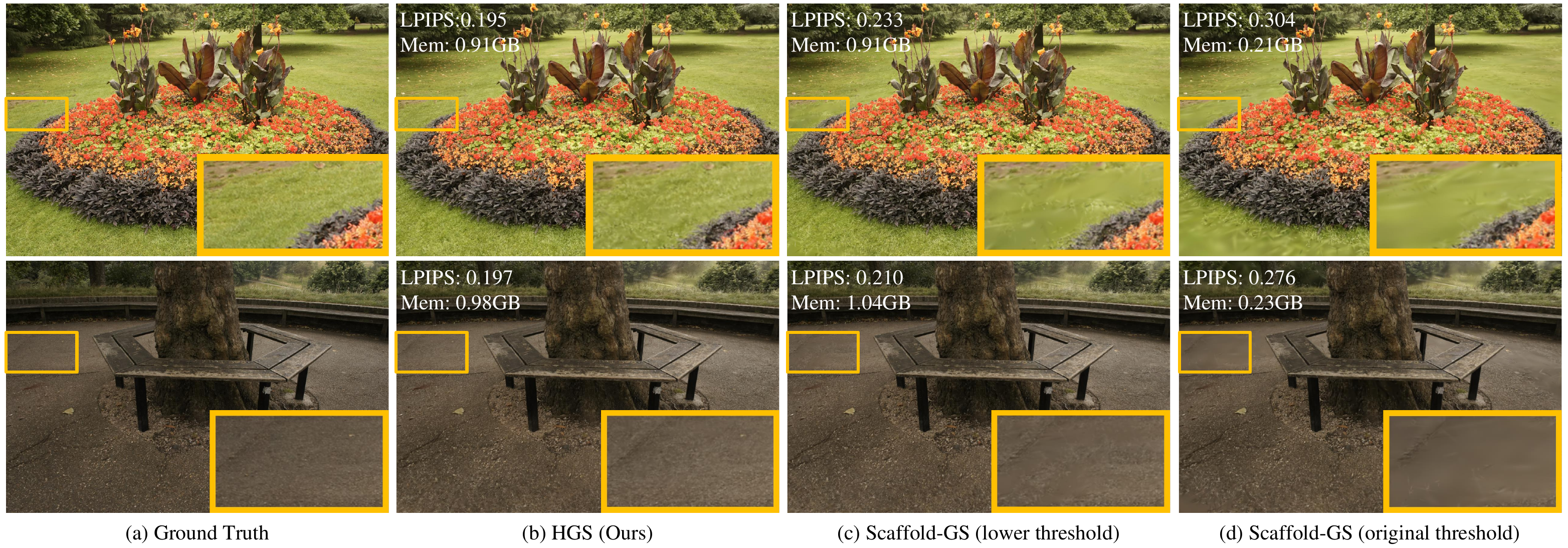}
    \caption{\textbf{Visual analysis of lowering $\tau_\text{grad}$ on the Mip-NeRF360 dataset~\cite{barron2022mip}.}}
    \label{fig:grad}
\end{figure*}

We conduct experiments on Mip-NeRF360 dataset~\cite{barron2022mip} by lowering $\tau_\text{grad}$ from 2.0e-4 to 7.0e-5 for Scaffold-GS to make the final optimized number of Gaussian points comparable to our HGS. The quantitative and qualitative results are shown in Table~\ref{tab:analysis_growing} and Figure~\ref{fig:grad}, respectively. 
We observe that: 1) Lowering $\tau_\text{grad}$ can boost rendering performance for Scaffold-GS. However, the improved performance still falls behind ours. 2) Lowering $\tau_\text{grad}$ still cannot reduce artifacts effectively, as shown in the yellow boxes of Figure~\ref{fig:grad}. This is because lowering $\tau_\text{grad}$ tends to grow unnecessary Gaussians in areas where Gaussian points are already dense, while regions with poor reconstruction are largely unaffected. In contrast, our HGS aims to identify hard Gaussians in poorly reconstructed areas. By focusing on growing and optimizing these Gaussians, our HGS can improve rendering quality more effectively.

\begin{table}[h]
    \centering
    \resizebox{\linewidth}{!}{

\begin{tabular}{l|c|ccc|c}
\toprule
& $\tau_\text{grad}$ & {PSNR $\uparrow$} & {SSIM $\uparrow$} & {LPIPS $\downarrow$} & Memory \\
\midrule
\multirow{2}{*}{Scaffold-GS~\cite{lu2024scaffold}} & 2.0e-4 & 27.98 & 0.825 & 0.208 & \bf 0.17GB \\
 & 7.0e-5 & \underline{28.27} & \underline{0.835} & \underline{0.174} & 0.64GB \\
\midrule
\textbf{HGS (Ours)} & 2.0e-4 & \bf 28.30 & \bf 0.837 & \bf 0.166 & \underline{0.57GB} \\
\bottomrule
\end{tabular}

}
    \caption{\textbf{Impact of lowering $\tau_\text{grad}$ on the Mip-NeRF360 dataset~\cite{knapitsch2017tanks}.}}
    \label{tab:analysis_growing}
\end{table}

\section{Implementation Details}
\label{sec:implementation}

Following previous practices~\cite{kerbl20233d,yu2024mip,zhang2024pixel}, indoor scenes are downsampled by a factor of 2 and outdoor scenes by 4 for the Mip-NeRF360 dataset~\cite{barron2022mip}. Note that, we achieve this by adopting the downsampling implementation provided by the 3DGS repository instead of directly using the downsampled images provided by Mip-NeRF360 dataset. All scenes on the Tanks\&Temples dataset~\cite{knapitsch2017tanks} are downsampled by a factor of 2 while the scenes on the Deep Blending dataset~\cite{hedman2018deep} use full-resolution images. In addition, we also evaluate our methods using the downsamled images provided by the Mip-NeRF360 dataset, the results are reported in Table~\ref{tab:mip360_i}. Our HGS still significantly outperforms SOTA methods in terms of PSNR and LPIPS metrics. Effi-HGS achieves competitive performance with SOTA methods.

\begin{table}[h]
    \centering
    \resizebox{0.9\linewidth}{!}{\begin{tabular}{l|ccc}
\toprule
& PSNR \(\uparrow\) & SSIM \(\uparrow\) & LPIPS \(\downarrow\) \\
\midrule
3DGS~\cite{kerbl20233d} & 27.41 & 0.813 & 0.218 \\
Pixel-GS~\cite{zhang2024pixel} & 27.53 & 0.822 & \cellcolor{tabthird}0.191 \\
Mip-Splatting~\cite{yu2024mip} & 27.65 & \cellcolor{tabfirst}0.828 & \cellcolor{tabsecond}0.188 \\
Scaffold-GS~\cite{lu2024scaffold} & \cellcolor{tabthird}27.67 & 0.812 & 0.223 \\
\midrule
\textbf{Effi-HGS (Ours)} & \cellcolor{tabsecond}27.92 & \cellcolor{tabthird}0.823 & 0.203 \\
\textbf{HGS (Ours)} & \cellcolor{tabfirst}28.06 & \cellcolor{tabsecond}0.827 & \cellcolor{tabfirst}0.182 \\
\bottomrule
\end{tabular}}
    \caption{\textbf{Quantitative results on the Mip-NeRF360 dataset~\cite{barron2022mip} using its provided downsampled images.}}
    \label{tab:mip360_i}
\end{table}

\section{Additional Ablation Study}
\label{sec:ablation_db}

We show additional ablation study results in Table~\ref{tab:ablation_db}. These results further support our analysis made in Section~\ref{sec:analysis}.

\begin{table}[h]
    \centering
    \resizebox{\linewidth}{!}{
\begin{tabular}{l|ccc|ccc}
\toprule
& OG & PGHGS & REHGS & {PSNR $\uparrow$} & {SSIM $\uparrow$} & {LPIPS $\downarrow$} \\
\midrule
Baseline & \checkmark & & & 30.23 & 0.907 & 0.245 \\
\midrule
Model-A & \checkmark & \checkmark & & 30.33 & \bf 0.908 & \bf 0.225 \\
Model-B &  \checkmark & & \checkmark & 30.32 & 0.907 & 0.245 \\
HGS & \checkmark & \checkmark & \checkmark & \bf 30.40 & \bf 0.908 & \bf 0.225  \\
\bottomrule
\end{tabular}}
    \caption{\textbf{Ablation study on Deep Blending~\cite{knapitsch2017tanks}.} OG: Original Growing criterion. PGHGS: Positional Gradient driven HGS. REHGS: Rendering Error guided HGS.}
    \label{tab:ablation_db}
\end{table}

\section{Additional Results}
\label{sec:additional}

In this section, we provide more qualitative and quantitative results on Mip-NeRF360~\cite{barron2022mip}, Tanks\&Temples~\cite{knapitsch2017tanks} and Deep Blending~\cite{hedman2018deep}.

\begin{figure*}[h]
    \centering
    \includegraphics[width=\linewidth]{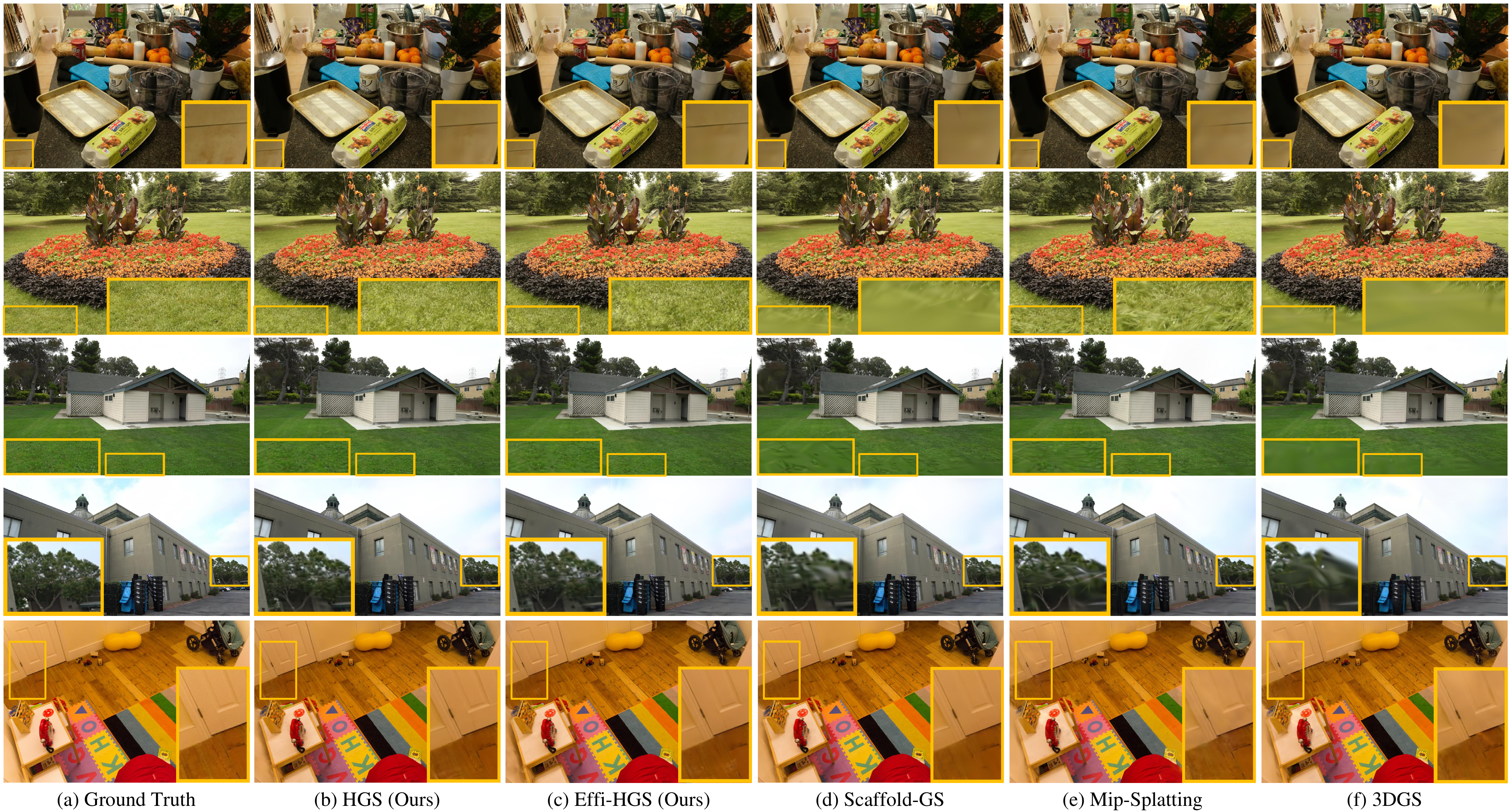}
    \caption{\textbf{More qualitative comparisons on three datasets~\cite{barron2022mip,knapitsch2017tanks,hedman2018deep}.} Yellow boxes show challenging areas. Our methods can reconstruct these areas well while other methods fail.}
    \label{fig:vis_comp_supp}
\end{figure*}

Tables~\ref{tab:mip360_perscene}, \ref{tab:tnt_perscene} and \ref{tab:db_perscene} show per-scene quantitative results on the Mip-NeRF360, Tanks\&Temples and Deep Blending, respectively. Our HGS achieves the best performance on most scenes of these datasets. Meanwhile, our Effi-HGS almost achieves the second-best performance. These detailed per-scene results further demonstrate the effectiveness of our proposed methods. Moreover, the results in Figure~\ref{fig:vis_comp_supp} show that our methods can reconstruct challenging details to reduce artifacts. 

\begin{table*}[h]
    \centering
    \small
    \begin{tabular}{l|ccccc|cccc}
 & \multicolumn{9}{c}{\textbf{PSNR $\uparrow$}} \\
 & \scenename{bicycle}  & \scenename{flowers}  & \scenename{garden}  & \scenename{stump}  & \scenename{treehill}  & \scenename{room}  & \scenename{counter}  & \scenename{kitchen} & \scenename{bonsai} 
 \\ 
 \hline 
Plenoxels~\cite{fridovich2022plenoxels} & 21.91 & 20.10 & 23.49 & 20.66 & 22.25 & 27.59 & 23.62 & 23.42 & 24.67 
\\
INGP~\cite{muller2022instant} & 22.19 & 20.35 & 24.60 & 23.63 & 22.36 & 29.27 & 26.44 & 28.55 & 30.34 
\\
Mip-NeRF360~\cite{barron2022mip} & 24.37 & 21.73 & 26.98 & 26.40 & 22.87 & 31.63 & 29.55 & \cellcolor{tabfirst}32.23 & \cellcolor{tabsecond}33.46 
\\
3DGS~\cite{kerbl20233d} & 25.59 & 21.84 & 27.74 & 26.92 & 22.80 & 31.63 & 29.11 & 31.45 & 32.30 
\\
Pixel-GS~\cite{zhang2024pixel} & \cellcolor{tabthird}25.74 & \cellcolor{tabthird}21.93 & 27.88 & 27.17 & 22.55 & 31.84 & 29.26 & 31.81 & 32.64 
\\
Mip-Splatting~\cite{yu2024mip} & \cellcolor{tabfirst}25.94 & \cellcolor{tabfirst}22.00 & \cellcolor{tabthird}27.97 & \cellcolor{tabthird}27.21 & 22.61 & 31.81 & 29.36 & \cellcolor{tabthird}31.93 & 32.50 
\\
Scaffold-GS~\cite{lu2024scaffold} & 25.68 & 21.71 & 27.83 & 26.83 & \cellcolor{tabfirst}23.45 & \cellcolor{tabthird}32.10 & \cellcolor{tabthird}29.66 & 31.80 & 32.74 
\\
\hline
\textbf{Effi-HGS (Ours)} & \cellcolor{tabsecond}25.88 & 21.66 & \cellcolor{tabsecond}28.06 & \cellcolor{tabsecond}27.23 & \cellcolor{tabsecond}23.14 & \cellcolor{tabsecond}32.38 & \cellcolor{tabsecond}29.88 & \cellcolor{tabsecond}31.94 & \cellcolor{tabthird}33.13 
\\
\textbf{HGS (Ours)} & \cellcolor{tabfirst}25.94 & \cellcolor{tabsecond}21.94 & \cellcolor{tabfirst}28.33 & \cellcolor{tabfirst}27.28 & \cellcolor{tabthird}23.06 & \cellcolor{tabfirst}32.61 & \cellcolor{tabfirst}30.26 & 31.63 & \cellcolor{tabfirst}33.64 
\\

\multicolumn{9}{c}{} \\
 & \multicolumn{9}{c}{\textbf{SSIM $\uparrow$}} \\
 & \scenename{bicycle}  & \scenename{flowers}  & \scenename{garden}  & \scenename{stump}  & \scenename{treehill}  & \scenename{room}  & \scenename{counter}  & \scenename{kitchen} & \scenename{bonsai} 
\\ 
\hline 
Plenoxels~\cite{fridovich2022plenoxels} & 0.496 & 0.431 & 0.606 & 0.523 & 0.509 & 0.842 & 0.759 & 0.648 & 0.814 
\\
INGP~\cite{muller2022instant} & 0.491 & 0.450 & 0.649 & 0.574 & 0.518 & 0.855 & 0.798 & 0.818 & 0.890 
\\
Mip-NeRF360~\cite{barron2022mip} & 0.685 & 0.583 & 0.813 & 0.744 & 0.632 & 0.913 & 0.894 & 0.920 & 0.941 
\\
3DGS~\cite{kerbl20233d} & 0.777 & 0.621 & 0.874 & 0.784 & \cellcolor{tabthird}0.653 & 0.927 & 0.914 & 0.932 & 0.947 
\\
Pixel-GS~\cite{zhang2024pixel} & \cellcolor{tabthird}0.792 & \cellcolor{tabsecond}0.653 & \cellcolor{tabsecond}0.878 & \cellcolor{tabsecond}0.797 & \cellcolor{tabthird}0.653 & 0.930 & \cellcolor{tabthird}0.920 & \cellcolor{tabfirst}0.936 & \cellcolor{tabthird}0.951 
\\
Mip-Splatting~\cite{yu2024mip} & \cellcolor{tabfirst}0.804 & \cellcolor{tabfirst}0.655 & \cellcolor{tabfirst}0.884 & \cellcolor{tabfirst}0.802 & \cellcolor{tabsecond}0.656 & \cellcolor{tabthird}0.933 & \cellcolor{tabthird}0.920 & \cellcolor{tabfirst}0.936 & \cellcolor{tabthird}0.951 
\\
Scaffold-GS~\cite{lu2024scaffold} & 0.773 & 0.611 & 0.869 & 0.777 & \cellcolor{tabfirst}0.661 & 0.931 & 0.918 & \cellcolor{tabthird}0.933 & 0.948 
\\
\hline
\textbf{Effi-HGS (Ours)} & 0.787 & 0.631 & \cellcolor{tabthird}0.877 & \cellcolor{tabthird}0.793 & 0.644 & \cellcolor{tabsecond}0.934 & \cellcolor{tabsecond}0.922 & \cellcolor{tabsecond}0.934 & \cellcolor{tabsecond}0.952 
\\
\textbf{HGS (Ours)} & \cellcolor{tabsecond}0.797 & \cellcolor{tabthird}0.646 & \cellcolor{tabfirst}0.884 & \cellcolor{tabsecond}0.797 & 0.652 & \cellcolor{tabfirst}0.938 & \cellcolor{tabfirst}0.928 & 0.932 & \cellcolor{tabfirst}0.956 
\\

\multicolumn{9}{c}{} \\
 & \multicolumn{9}{c}{\textbf{LPIPS $\downarrow$}} \\
 & \scenename{bicycle}  & \scenename{flowers}  & \scenename{garden}  & \scenename{stump}  & \scenename{treehill}  & \scenename{room}  & \scenename{counter}  & \scenename{kitchen} & \scenename{bonsai} 
\\ 
\hline 
Plenoxels~\cite{fridovich2022plenoxels} & 0.506 & 0.521 & 0.386 & 0.503 & 0.540 & 0.419 & 0.441 & 0.447 & 0.398 
\\
INGP~\cite{muller2022instant} & 0.487 & 0.481 & 0.312 & 0.450 & 0.489 & 0.301 & 0.342 & 0.254 & 0.227  
\\
Mip-NeRF360~\cite{barron2022mip} & 0.301 & 0.344 & 0.170 & 0.261 & 0.339 & 0.211 & 0.204 & 0.127 & 0.176 
\\
3DGS~\cite{kerbl20233d} & 0.205 & 0.330 & 0.103 & 0.207 & 0.318 & 0.192 & 0.179 & 0.114 & 0.174 
\\
Pixel-GS~\cite{zhang2024pixel} & \cellcolor{tabthird}0.173 & \cellcolor{tabsecond}0.254 & \cellcolor{tabthird}0.093 & \cellcolor{tabfirst}0.180 & \cellcolor{tabthird}0.269 & 0.183 & \cellcolor{tabsecond}0.163 & \cellcolor{tabfirst}0.106 & \cellcolor{tabthird}0.161 
\\
Mip-Splatting~\cite{yu2024mip} & \cellcolor{tabfirst}0.162 & 0.266 & \cellcolor{tabsecond}0.090 & \cellcolor{tabsecond}0.181 & 0.270 & \cellcolor{tabthird}0.175 & 0.165 & \cellcolor{tabsecond}0.107 & \cellcolor{tabsecond}0.157 
\\
Scaffold-GS~\cite{lu2024scaffold} & 0.223 & 0.340 & 0.111 & 0.229 & 0.315 & 0.183 & 0.177 & 0.115 & 0.173 
\\
\hline
\textbf{Effi-HGS (Ours)} & 0.186 & \cellcolor{tabthird}0.265 & 0.097 & 0.190 & \cellcolor{tabsecond}0.265 & \cellcolor{tabsecond}0.174 & \cellcolor{tabthird}0.164 & \cellcolor{tabthird}0.110 & 0.162 
\\
\textbf{HGS (Ours)} & \cellcolor{tabsecond}0.165 & \cellcolor{tabfirst}0.247 & \cellcolor{tabfirst}0.086 & \cellcolor{tabthird}0.183 & \cellcolor{tabfirst}0.242 & \cellcolor{tabfirst}0.163 & \cellcolor{tabfirst}0.149 & 0.112 & \cellcolor{tabfirst}0.148 
\\

\multicolumn{9}{c}{}
\end{tabular}

    \caption{\textbf{Per-scene quantitative results on the Mip-NeRF 360 dataset~\cite{barron2022mip}.}
    }
    \label{tab:mip360_perscene}
\end{table*}

\begin{table*}[h]
    \centering
    \small
    \begin{tabular}{l|cccccc}
 & \multicolumn{6}{c}{\textbf{PSNR $\uparrow$}} \\
 & \scenename{Barn}  & \scenename{Caterpillar}  & \scenename{Courthouse}  & \scenename{Ignatius}  & \scenename{Meetingroom}  & \scenename{Truck} 
 \\ 
 \hline 
3DGS~\cite{kerbl20233d} & 28.79 & 24.08 & 22.57 & 22.31 & 26.04 & 25.61 
\\
Pixel-GS~\cite{zhang2024pixel} & 29.38 & 24.42 & 22.80 & \cellcolor{tabthird}22.66 & 26.07 & 25.71 
\\
Mip-Splatting~\cite{yu2024mip} & \cellcolor{tabthird}29.60 & 24.17 & 21.99 & 22.26 & 26.13 & 25.97 
\\
Scaffold-GS~\cite{lu2024scaffold} & 29.57 & \cellcolor{tabthird}24.60 & \cellcolor{tabthird}23.57 & \cellcolor{tabsecond}23.38 & \cellcolor{tabthird}27.12 & \cellcolor{tabthird}26.18 
\\
\hline
\textbf{Effi-HGS (Ours)} & \cellcolor{tabsecond}29.92 & \cellcolor{tabsecond}24.86 & \cellcolor{tabsecond}23.97 & \cellcolor{tabfirst}23.78 & \cellcolor{tabsecond}27.31 & \cellcolor{tabsecond}26.30 
\\
\textbf{HGS (Ours)} & \cellcolor{tabfirst}30.49 & \cellcolor{tabfirst}25.53 & \cellcolor{tabfirst}24.26 & \cellcolor{tabfirst}23.78 & \cellcolor{tabfirst}27.63 & \cellcolor{tabfirst}26.46 
\\

\multicolumn{6}{c}{} \\
 & \multicolumn{6}{c}{\textbf{SSIM $\uparrow$}} \\
 & \scenename{Barn}  & \scenename{Caterpillar}  & \scenename{Courthouse}  & \scenename{Ignatius}  & \scenename{Meetingroom}  & \scenename{Truck} 
\\ 
\hline 
3DGS~\cite{kerbl20233d} & 0.879 & 0.822 & 0.797 & 0.824 & 0.890 & 0.886 
\\
Pixel-GS~\cite{zhang2024pixel} & 0.897 & \cellcolor{tabthird}0.844 & 0.803 & \cellcolor{tabthird}0.833 & 0.890 & 0.891 
\\
Mip-Splatting~\cite{yu2024mip} & \cellcolor{tabthird}0.900 & 0.834 & 0.783 & 0.831 & 0.896 & \cellcolor{tabsecond}0.899 
\\
Scaffold-GS~\cite{lu2024scaffold} & 0.889 & 0.828 & \cellcolor{tabthird}0.821 & \cellcolor{tabthird}0.833 & \cellcolor{tabthird}0.902 & 0.891 
\\
\hline
\textbf{Effi-HGS (Ours)} & \cellcolor{tabsecond}0.901 & \cellcolor{tabsecond}0.845 & \cellcolor{tabsecond}0.834 & \cellcolor{tabsecond}0.841 & \cellcolor{tabsecond}0.906 & \cellcolor{tabthird}0.895
\\
\textbf{HGS (Ours)} & \cellcolor{tabfirst}0.914 & \cellcolor{tabfirst}0.867 & \cellcolor{tabfirst}0.841 & \cellcolor{tabfirst}0.845 & \cellcolor{tabfirst}0.911 & \cellcolor{tabfirst}0.900 
\\

\multicolumn{6}{c}{} \\
 & \multicolumn{6}{c}{\textbf{LPIPS $\downarrow$}} \\
 & \scenename{Barn}  & \scenename{Caterpillar}  & \scenename{Courthouse}  & \scenename{Ignatius}  & \scenename{Meetingroom}  & \scenename{Truck} 
\\ 
\hline 
3DGS~\cite{kerbl20233d} & 0.170 & 0.198 & 0.245 & 0.159 & 0.160 & 0.137 
\\
Pixel-GS~\cite{zhang2024pixel} & \cellcolor{tabthird}0.133 & \cellcolor{tabsecond}0.161 & 0.236 & \cellcolor{tabthird}0.142 & 0.156 & \cellcolor{tabsecond}0.112
\\
Mip-Splatting~\cite{yu2024mip} & 0.142 & 0.181 & 0.261 & 0.149 & \cellcolor{tabthird}0.150 & 0.116 
\\
Scaffold-GS~\cite{lu2024scaffold} & 0.156 & 0.200 & \cellcolor{tabthird}0.215 & 0.156 & 0.159 & 0.132 
\\
\hline
\textbf{Effi-HGS (Ours)} & \cellcolor{tabsecond}0.128 & \cellcolor{tabthird}0.170 & \cellcolor{tabsecond}0.191 & \cellcolor{tabsecond}0.136 & \cellcolor{tabsecond}0.146 & \cellcolor{tabthird}0.114 
\\
\textbf{HGS (Ours)} & \cellcolor{tabfirst}0.100 & \cellcolor{tabfirst}0.134 & \cellcolor{tabfirst}0.180 & \cellcolor{tabfirst}0.122 & \cellcolor{tabfirst}0.129 & \cellcolor{tabfirst}0.092 
\\

\multicolumn{6}{c}{}
\end{tabular}

    \caption{\textbf{Per-scene quantitative results on the Tanks\&Temples dataset~\cite{knapitsch2017tanks}.}}
    \label{tab:tnt_perscene}
\end{table*}

\begin{table*}[h]
    \centering
    \small
    
\begin{tabular}{l|cc|cc|cc}
 & \multicolumn{2}{c|}{\textbf{PSNR $\uparrow$}} & \multicolumn{2}{c|}{\textbf{SSIM $\uparrow$}} & \multicolumn{2}{c}{\textbf{LPIPS $\downarrow$}} \\
 & \scenename{Dr Johnson}  & \scenename{Playroom} & \scenename{Dr Johnson}  & \scenename{Playroom} & \scenename{Dr Johnson}  & \scenename{Playroom} 
 \\ 
 \hline 
Plenoxels~\cite{fridovich2022plenoxels} & 23.14 & 22.98 & 0.787 & 0.802 & 0.521 & 0.499 
\\
INGP~\cite{muller2022instant} & 27.75 & 19.48 & 0.839 & 0.754 & 0.381 & 0.465 
\\
Mip-NeRF360~\cite{barron2022mip} & 29.14 & 29.66 & \cellcolor{tabthird}0.901 & \cellcolor{tabthird}0.900 & \cellcolor{tabthird}0.237 & 0.252 
\\
3DGS~\cite{kerbl20233d} & 29.05 & \cellcolor{tabthird}29.88 & 0.898 & \cellcolor{tabthird}0.900 & 0.247 & 0.246  
\\
Pixel-GS~\cite{zhang2024pixel} & 28.02 & 29.74 & 0.885 & 0.899 & 0.257 & 0.244 
\\
Mip-Splatting~\cite{yu2024mip} & 28.75 & 29.74 & 0.899 & \cellcolor{tabsecond}0.907 & 0.243 & \cellcolor{tabsecond}0.236 
\\
Scaffold-GS~\cite{lu2024scaffold} & \cellcolor{tabthird}29.74 & 30.73 & \cellcolor{tabsecond}0.906 & \cellcolor{tabfirst}0.908 & 0.243 & 0.247 
\\
\hline
\textbf{Effi-HGS (Ours)} & \cellcolor{tabsecond}29.79 & \cellcolor{tabsecond}30.87 & \cellcolor{tabfirst}0.908 & \cellcolor{tabfirst}0.908 & \cellcolor{tabsecond}0.235 & \cellcolor{tabthird}0.240 
\\
\textbf{HGS (Ours)} & \cellcolor{tabfirst}29.92 & \cellcolor{tabfirst}30.89 & \cellcolor{tabfirst}0.908 & \cellcolor{tabfirst}0.908 & \cellcolor{tabfirst}0.225 & \cellcolor{tabfirst}0.225 
\\

\multicolumn{6}{c}{}
\end{tabular}

    \caption{\textbf{Per-scene quantitative results on the Deep Blending dataset~\cite{hedman2018deep}.}}
    \label{tab:db_perscene}
\end{table*}

\section{Impact for Anti-aliasing Ability}
\label{sec:anti-aliasing}

We further investigate the impact of our method for anti-aliasing ability of Mip-Splatting~\cite{yu2024mip} on Mip-NeRF360 dataset~\cite{barron2022mip}. Results are reported in Table~\ref{tab:mip360_stmt}. We observe that our method can further improve the anti-aliasing ability for Mip-Splatting, further demonstrating its potential generalization ability to other Gaussian splatting related tasks.

\begin{table*}[h]
    \centering
    \resizebox{\linewidth}{!}{
\begin{tabular}{@{}l@{\,\,}|ccccc|ccccc|ccccc}
& \multicolumn{5}{c|}{PSNR $\uparrow$} & \multicolumn{5}{c|}{SSIM $\uparrow$} & \multicolumn{5}{c}{LPIPS $\downarrow$}  \\
& $1 \times$ Res. & $2 \times$ Res. & $4 \times$ Res. & $8 \times$ Res. & Avg. & $1 \times$ Res. & $2 \times$ Res. & $4 \times$ Res. & $8 \times$ Res. & Avg. & $1 \times$ Res. & $2 \times$ Res. & $4 \times$ Res. & $8 \times$ Res. & Avg.  \\ 
\hline    
INGP~\cite{muller2022instant} & 26.79  & 24.76 & \cellcolor{tabthird}24.27 & 24.27 & 25.02 & 0.746 & 0.639 & 0.626 & 0.698 & 0.677 & 0.239 & 0.367 & 0.445 & 0.475 & 0.382 
\\
Mip-NeRF360~\cite{barron2022mip} & 29.26 & 25.18 & 24.16 & 24.10 & 25.67 & 0.860 & 0.727 & 0.670 & \cellcolor{tabthird}0.706 & 0.741 & 0.122 & 0.260 & 0.370 & 0.428 & 0.295 
\\
zip-NeRF~\cite{barron2023zip} & \cellcolor{tabfirst}29.66 & 23.27 & 20.87 & 20.27  & 23.52 & 0.875 & 0.696 & 0.565 & 0.559 & 0.674 & 0.097 & 0.257 & 0.421 & 0.494 & 0.318 
\\
3DGS~\cite{kerbl20233d} & 29.19 & 23.50  & 20.71 & 19.59 & 23.25 & \cellcolor{tabthird}0.880 & 0.740 & 0.619 & 0.619 & 0.715 & 0.107 & 0.243 & 0.394 & 0.476 & 0.305 \\
\hline
Mip-Splatting & 29.39 & \cellcolor{tabthird}27.48 & \cellcolor{tabsecond}26.54 & \cellcolor{tabthird}26.29 & \cellcolor{tabthird}27.43 & \cellcolor{tabsecond}0.890 & \cellcolor{tabsecond}0.815 & \cellcolor{tabsecond}0.759 & \cellcolor{tabfirst}0.768  & \cellcolor{tabsecond}0.808 & \cellcolor{tabsecond}0.092 & \cellcolor{tabsecond}0.188 & \cellcolor{tabsecond}0.292 & \cellcolor{tabsecond}0.385 & \cellcolor{tabsecond}0.239 \\
Mip-Splatting* & \cellcolor{tabthird}29.51  & \cellcolor{tabsecond}27.49 & \cellcolor{tabsecond}26.54 & \cellcolor{tabsecond}26.30 & \cellcolor{tabsecond}27.46 & \cellcolor{tabsecond}0.890 & \cellcolor{tabthird}0.814 & \cellcolor{tabthird}0.758 & \cellcolor{tabsecond}0.767 & \cellcolor{tabthird}0.807 & \cellcolor{tabthird}0.093 & \cellcolor{tabthird}0.192 & \cellcolor{tabthird}0.295 & \cellcolor{tabthird}0.388 & \cellcolor{tabthird}0.242 \\
\textbf{Mip-Splatting* + HGS (Ours)} & \cellcolor{tabsecond}29.58 & \cellcolor{tabfirst}27.54 & \cellcolor{tabfirst}26.57 & \cellcolor{tabfirst}26.31 & \cellcolor{tabfirst}27.50 & \cellcolor{tabfirst}0.891 & \cellcolor{tabfirst}0.818 & \cellcolor{tabfirst}0.762 & \cellcolor{tabsecond}0.767 & \cellcolor{tabfirst}0.809 & \cellcolor{tabfirst}0.088 & \cellcolor{tabfirst}0.183 & \cellcolor{tabfirst}0.287 & \cellcolor{tabfirst}0.383 & \cellcolor{tabfirst}0.235  
\end{tabular}}
    \caption{\textbf{Single-scale training and multi-scale testing on the Mip-NeRF360 dataset~\cite{barron2022mip}.} All methods are trained on the smallest scale (1$\times$) and evaluated across four scales (1$\times$, 2$\times$, 4$\times$, and 8$\times$), with evaluations at higher sampling rates simulating zoom-in effects. Mip-Splatting* means it is combined with opacity correction~\cite{bulo2024revising}.}
    \label{tab:mip360_stmt}
\end{table*}

\section{Discussions and Limitations}
\label{sec:discussion}

Like previous methods~\cite{kerbl20233d,lu2024scaffold}, our methods rely on sparse point clouds from SfM to initialize Gaussian points and then grow new Gaussians from existing Gaussians gradually. This may require a longer growth process for distant regions that are not reconstructed well. One possible way to accelerate this process is to introduce structured geometry priors~\cite{xu2020planar,xu2022multi,sinha2009piecewise} to guide the growing for these areas. Additionally, since our methods grow more Gaussians to boost rendering quality, their efficiency decreases slightly. Designing
 an efficient coarse-to-fine/multi-scale training strategy~\cite{girish2023eagles,xu2019multi,yan2024multi} is an avenue for future work. 

\end{document}